# Qualitative propagation and scenario-based approaches to explanation of probabilistic reasoning*

**Max Henrion**[1, 2] and **Marek J. Druzdzel**[1]

[1]Department of Engineering and Public Policy
Carnegie Mellon University
Pittsburgh, Pa 15213, USA

[2]Rockwell International Science Center
Palo Alto Laboratory, 444 High Street, #400
Palo Alto, Ca 94301, USA

## Abstract

Comprehensible explanations of probabilistic reasoning are a prerequisite for wider acceptance of Bayesian methods in expert systems and decision support systems. A study of human reasoning under uncertainty suggests two different strategies for explaining probabilistic reasoning specially attuned to human thinking: The first, *qualitative belief propagation*, traces the qualitative effect of evidence through a belief network from one variable to the next. This propagation algorithm is an alternative to the graph reduction algorithms of Wellman (1988) for inference in qualitative probabilistic networks. It is based on a qualitative analysis of intercausal reasoning, which is a generalization of Pearl's "explaining away", and an alternative to Wellman's definition of qualitative synergy. The other, *Scenario-based reasoning*, involves the generation of alternative causal "stories" accounting for the evidence. Comparing a few of the most probable scenarios provides an approximate way to explain the results of probabilistic reasoning. Both schemes employ causal as well as probabilistic knowledge. Probabilities may be presented as phrases and/or numbers. Users can control the style, abstraction and completeness of explanations.

## 1 Introduction

The developers of expert systems and decision support systems have long been aware of the importance of facilities to explain the computer-based reasoning to users as a prerequisite to their more widespread acceptance (e.g. Teach & Shortliffe, 1981). Unless users can come to understand the assumptions and reasoning of such systems, it is impossible to develop the kind of human-machine collaboration that is the basis for successful use of such systems. For explanations to be effective, their form and content must be carefully matched to the users' competence, knowledge, and styles of reasoning.

The approach that underlies the classic "expert systems" paradigm is to employ computer representations and inference mechanisms intended to emulate human reasoning. To the extent that this emulation is successful, the computer-based reasoning ought to seem familiar to people and so relatively easy to explain. While there is ample evidence that normatively appealing probabilistic and decision theoretic schemes are poor models of human reasoning under uncertainty (e.g. Kahneman *et al.* 1982), there is surprisingly little experimental evidence that the rule-based alternatives, such as certainty factors or fuzzy logic, are any better as descriptive models. And even if successful descriptively, the emulative approach would merely reproduce the documented deficiencies of our intuitive reasoning rather than complement and enhance it. Are we forced to choose between the unreliable and the inexplicable? Our approach to this dilemma is to explore whether in fact it may be possible to explicate probabilistic reasoning in ways better attuned to human thinking.

Only recently has much attention begun to be paid to the automatic generation of comprehensible explanations for probabilistic and decision analytic schemes. Horvitz *et al.* (1986) present a system which can explain its recommendations about what test to perform to gather diagnostic evidence. Langlotz *et al.* (1986) present a scheme for quantitative analysis of decision trees, which explains qualitatively how one decision may outweigh another in terms of expected utility. Klein (1990) presents a scheme for qualitatively explaining the implications of hierarchical additive value functions. Elsaesser (1988) provides some empirical evidence on the efficacy of explanations of simple Bayesian inference, with one variable and one observation. Strat's (1988) system explains the dynamics of Dempster-Shafer reasoning based on sensitivity analysis, with interesting implications for probabilistic schemes. Sember and Zukerman's (1989) scheme generates *micro* explanations, that is local propagation of evidence between neighbouring variables in a belief net.

* This work was supported by the National Science Foundation under grant IRI-8807061 to Carnegie Mellon and by the Rockwell International Science Center.

Our focus here is on approaches for generating *macro* explanations, intended to explain probabilistic reasoning over larger networks. We wish to avoid dogmatism about what kinds of explanation scheme will be most effective, but rather explore a variety of approaches, including graphical, numerical, and linguistic representations. We are interested in both quantitative and qualitative forms of explanation in various combinations. This paper gives an account of several of the key ideas that have emerged from our initial work.

Since our goal is to produce interpretations of probabilistic reasoning that are more compatible with human reasoning styles, we started out with an empirical study of human strategies for uncertain reasoning. This provided us with the inspiration for the design of two new and contrasting modes of explaining probabilistic reasoning, namely *qualitative belief propagation* and *scenario-based reasoning*.

It is useful to distinguish explanation as the communication of *static* knowledge or beliefs from explanation of the *dynamics*, of how beliefs are changed in the light of new evidence. Explanation of the statics, though relatively straightforward, is a prerequisite for explanation of the dynamics. Among the issues in static explanation we discuss are the use of belief nets, the use of linguistic phrases to express probabilities, and the importance of causal knowledge. Next we outline the use of qualitative belief propagation as a means of dynamic explanation. This includes an analysis of qualitative intercausal reasoning, generalizing Pearl's notion of "explaining away", with a theorem giving a precise characterization of when it applies. Finally, we describe a scenario-based approach to explanation. This is illustrated by explanations generated from our prototype implementation in Allegro Common Lisp, QIQ (Qualitative Interface to the Quantitative).

## 2 Human reasoning under uncertainty

The essence of effective explanation is to design its content and form to mesh with the knowledge and modes of thought of the person to whom you are explaining. Thus, producing good explanations of formal reasoning under uncertainty requires an understanding of the way people reason intuitively under uncertainty. There is a vast literature on human judgment under uncertainty for very simple inference problems, typically with a single hypothesis variable and a single observation (e.g. see Kahneman *et al.*, 1982 and Morgan & Henrion, 1990 for reviews), but relatively little is known about cognitive processes in more complex situations. To improve our insights into this and to seek inspiration for alternative approaches to explanation, we conducted a series of cognitive process-tracing studies. We recorded and analyzed verbal protocols from subjects asked to think aloud as they performed uncertain reasoning tasks (Druzdzel, 1989). Here is a sample task:

> *Harry is in the house of a new acquaintance and suddenly finds himself sneezing. This could be due to an incipient cold, or to an allergy attack brought on by a cat. Before he started sneezing he would have judged the cold and allergy both about equally unlikely.*
>
> *(a) Given he is sneezing, roughly what is the probability Harry is getting a cold?*
>
> *(b) Suppose Harry now notices small paw-prints on the furniture. How should this affect his degree of belief that he is getting a cold?*
>
> *(c) Suppose he then hears a barking of a small dog in the room next-door. How does this further affect his degree of belief that he is getting a cold?*

Most subjects were able to provide qualitative answers to these questions rather easily. In (a) they judged a cold was about as likely as not. In (b) that the paw-prints should decrease his belief in the cold, since the sneezing might be explained by a cat, suggested by the paw-prints. And in (c), that hearing the barking dog should increase belief in the cold again, since the dog provides an alternative explanation of the paw prints.

One unsurprising finding was that subjects generally used qualitative terms for probabilities, using quantitative terms almost not at all. This finding is consistent with previous studies of intuitive reasoning (e.g. Kuipers, Moskovitz & Kassirer, 1988). Another finding confirming previous work (e.g. Kahneman and Tversky, 1980) was the importance of causal reasoning in uncertain inference.

Less expected and of considerable interest in the current context, was evidence of two quite different strategies for plausible reasoning. One, which we call *qualitative belief propagation*, involves propagating the qualitative impact of evidence from event to event, following local causal and diagnostic relationships. For example, barking indicates the presence of the dog. The dog explains the pawmarks, which are then weaker evidence for the cat. Reduced belief in the cat, in turn reduces belief in the allergy. This is now less of an explanation for the sneezing and so requires an increased belief in the cold.

The other strategy, *scenario-based reasoning*, is quite different, and was more common in the protocols. The reasoner identifies one or more scenarios, that is consistent instantiations of the variables, forming a coherent, often causal, story, compatible with the known evidence. For example, Harry has a cold, which explains the sneezing; there is no cat, and so no allergy; the dog explains the paw-prints and barking. Subjects often appeared to develop one or more such quasi-deterministic scenarios. Figure 1 shows two such scenarios, as a subset of the event tree.

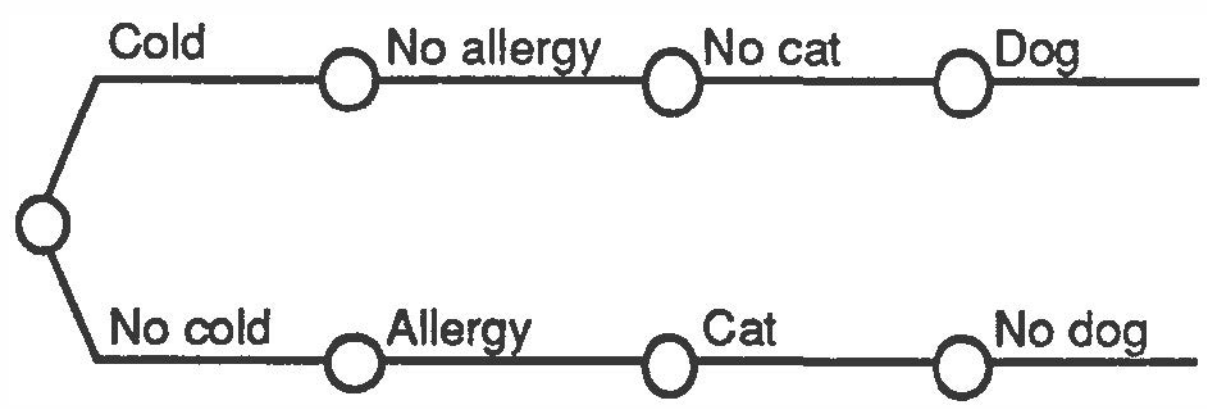

**Figure 1: Two scenarios for the sneeze problem**

The probability of some target event (e.g. the cold) can then be judged by the relative probability of the scenario(s) that contain(s) it. If one considered all possible scenarios, then this strategy is an exact algorithm for Bayesian reasoning. This is generally too much mental effort, but it can be a good approximation if one considers only the few most probable scenarios. On the other hand, if a likely scenario is ignored or its relative probability misestimated, it can lead to severe biases. This scenario-based reasoning appears related to explanation-based reasoning identified by a number of psychologists as strategies for complex reasoning tasks (e.g. Pennington & Hastie, 1988).

Both qualitative belief propagation and scenario-based reasoning can be seen as approximations of exact algorithms for probabilistic inference. They are suggestive of two quite different explanation strategies, which we will describe below.

## 3. Explanation of static probabilistic knowledge

### 3.1 Belief nets

By now the most familiar display of qualitative probabilistic information is the Bayesian belief net (and influence diagram), which provides a perspicuous display of purely qualitative beliefs about conditional dependence and independence. Figure 2 provides a belief network for probabilistic knowledge for the "sneeze" example.

The nodes depict the key variables. (NB, we use the abbreviated term "Cat" to mean "the presence of a cat in the vicinity", and so on.) As usual, the directed arcs depict dependences between them, (or more strictly the *absence* of arcs depicts *independence*). The same information could, of course, be represented in text as a list of the dependencies, such as,

```
The probability of sneezing is
affected by cold.
The probability of sneezing depends
on allergy.
```

and so on.

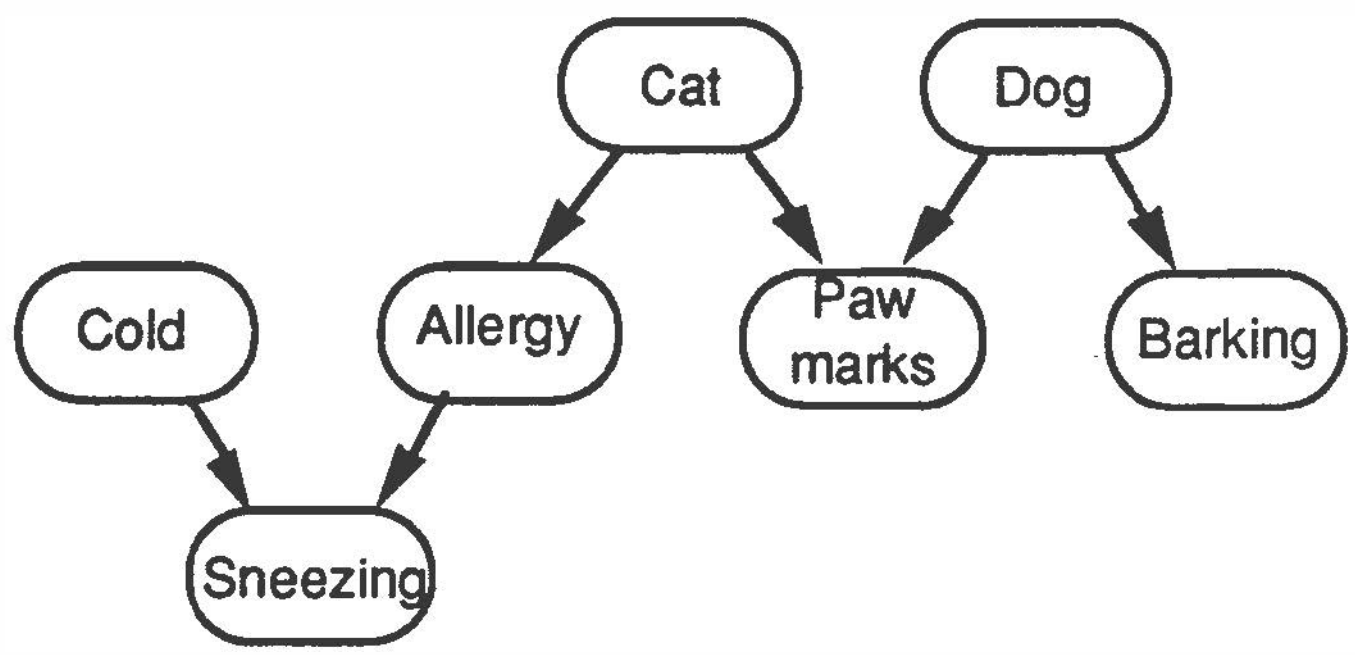

**Figure 2: A belief network for the "sneeze example"**

The improved perspicuity of the graphical representation in showing the locality of relationships is immediately clear. Although for some purposes the textual form is valuable, particularly for those not familiar with the belief network notation. To complete the static explanation we need to add the probabilities in some form.

### 3.2 Linguistic probabilities

One appealing approach to render numerical probabilities more digestible is to translate them into verbal phrases, such as "very likely" or "somewhat improbable". A considerable empirical literature reports people's interpretations of verbal probability phrases in terms of numerical probabilities or ranges. In general this research has found a degree of consistency in usage, at least in the ordering people assign to sets of such phrases (Budescu & Wallsten, 1985; Wallsten *et al*, 1986; Kong *et al.* 1986). But it has also found significant variability in interpretation between people, and considerable context dependence (Brun & Teigen, 1988). Nuclear safety engineers mean something quite different by "uncommon" than physicians. People interpret other people's use of phrases somewhat differently (and with wider range of uncertainty) from what they themselves claim to mean by the phrases (Wallsten *et al*; 1986); that is, they are sensitive to the variability among people. This suggests mappings from phrases to numbers, needed for encoding, should be somewhat different, with broader ranges than mappings from phrases to numbers, as used here for explanations.

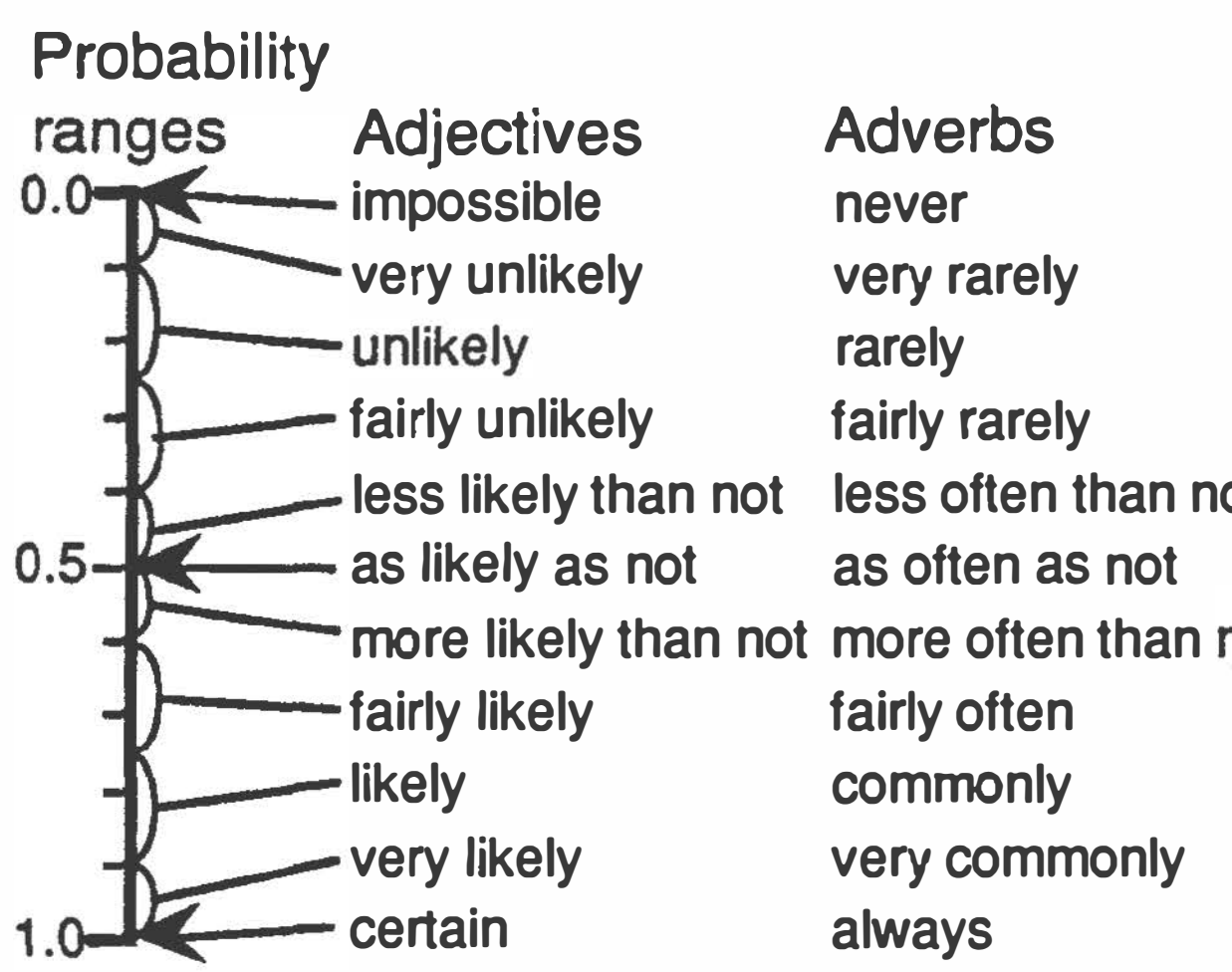

**Figure 3: Sample mappings from numerical probabilities to adjective and adverb phrases.**

To cope with differences in personal preference and the context-dependence of interpretations, our explanation system, QIQ, provides a variety of mappings, including two mappings from the literature (Wallsten *et al.* 1986; Kong *et al.* 1986), and our own synthesis from the literature, illustrated in figure 3. We have tried to use terms which minimize ambiguity and variability among people and contexts. Users can select from these mappings or provide their own. The context and interpretation may vary not only by domain, but even by event within a network. For example, "unlikely" may mean something quite different when applied to the chance of allergy to an antibiotic than the chance of dying in an operation. If desired, a different mapping may be used for each event and influence. However we expect a small number of mappings will be sufficient to cover the contexts for a given network.

Relevant phrases can be divided into *belief* phrases, such as "very probable" or "unlikely", and *frequency* phrases, such as "common" or "rare". Most come in both adjectival form, as above, and adverbial form, such as "probably" or "commonly". We have found it most natural to express marginal prior and posterior probabilities of events in terms of adjectival probabilities, for example,

```
Cold is very unlikely (p=0.08)
Cat is unlikely (p=0.1)
Dog is unlikely (p=0.1)
```

and to express conditional probabilities or causal strengths (see below) in terms of adverbial frequencies:

```
Cat commonly (p=0.8) causes allergy.
Dog as often as not (p=0.5) causes
barking.
```

The above examples are generated by QIQ as part of the static explanation of the sneeze belief network.

Due to variations between people and contexts, some vagueness in interpretation inevitably remains. To some this is part of the attraction of verbal phrases over numerical probabilities. Others may wish to see the numerical probability in addition to the verbal phrase, as in the examples above. This allows users to pay attention to whatever they find most helpful, and, with experience, perhaps to learn the mappings.

While most previous empirical work has examined absolute probabilities, one recent study (Elsaesser & Henrion, 1990) has examined mappings from *relative* probabilities or changes in probabilities to phrases such as "more likely than". They found that a fixed mapping to phrases from differences in probabilities provided a better model than ratios of probabilities or odds. These phrases are useful for comparing the probabilities of events or updates in degrees of belief, such as:

```
Cold is slightly less likely than
cat (0.08/0.10).
No cat is a great deal more likely
than cat (0.1/0.9).
```

### 3.3 Causal relationships

A key finding of the behavioral decision theory literature is the psychological importance of causal structure in uncertain reasoning (Tversky & Kahneman, 1980). People find it easier to reason from cause to effect than vice versa. As we mentioned, this was also apparent in our protocol studies. Some, notably Pearl (1988), have explicitly identified the directed arcs of belief nets with cause-effect relations. Others have argued that there is no inherent relationship with causality: After all, the arcs can be reversed simply by application of Bayes' rule, but causality cannot. But in any case, it is usually most natural to assess influences in causal direction. We view knowledge of causal relations as an important semantic enrichment to the pure belief net. It is not essential for Bayesian inference, but can be of great help in communicating with people, both for encoding expert opinion and for explanation.

QIQ can encode a cause-effect relationship as supplementary knowledge about each influence arc. This information is used in generating text descriptions of influences, for example:

```
Cat commonly (p=0.8) causes allergy.
Cat is the only cause of allergy.
```

The quantities described here are *causal strengths*, that is the probability that the specified precursor event, if present, is sufficient to cause the successor. If no other cause of the successor is present then the causal strength is the same as the conditional probability of the effect given the cause. This is the case with the link from cat to allergy, where no other cause is known (in this example). However, if other causes are possible, then the causal strengths may be different from the conditional probabilities. Causal strengths are an equivalent representation to the conditional probability representation, and each can be derived from the other.

The best known application of causal strengths is in the noisy-OR gate, which often arises in situations with multiple alternative causes of a common effect. Each link from cause to effect is characterized by its causal strength, the probability of the effect given only that cause is present. The condition it embodies is sometimes called *causal independence,* namely that the probability that each present cause is sufficient to produce the effect is independent of the presence or sufficiency of other causes. For example, we have,

```
Cold very commonly (p=0.9) causes
sneezing.
Allergy very commonly (p=0.9) causes
sneezing.
```

Causal independence can be expressed as:

```
Cold does not affect the tendency of
allergy to cause sneezing, and vice
versa.
```

In this case we also assume no *leaks* (Henrion, 1990), i.e. no "spontaneous" occurence of the effect in the absence of explicitly modelled causes:

```
There is no other cause of sneezing
than cold and allergy
```

In other cases we cannot rule out leaks, for example:

```
Cat as often as not (p=0.5) causes
paw marks.
Dog as often as not (p=0.5) causes
paw marks.
There are also other very unlikely
(p=0.1) causes of paw marks.
```

The latter assertion gives the *leak probability*, that is the probability of paw marks given no cat or dog. Note that the former two assertions do not give the simple conditional probability of paw marks given the cat (dog), but given also none of the "other causes" mentioned in the last assertion.

The entire static description of the belief net used in these examples is completed by the following assertions:

```
Dog as often as not (p=0.5) causes
barking.
Dog is the only cause of barking.
```

## 4 Qualitative belief propagation

The goal of qualitative belief propagation is to find the direction of the impact of an observed variable on the degree of belief in another variable, whether increased, decreased, or unchanged (+ - 0). Wellman (1988) presents a scheme for qualitative probabilistic networks (QPNs) which provides an appealing formal basis for this task for arbitrary belief nets. Wellman's scheme uses an inference algorithm for QPNs using arc reversal and graph reduction, modelled on Shachter's algorithms for inference in quantitative belief networks. However, human qualitative belief propagation appears to trace the impact of evidence locally from node to node, which seems more reminiscent of the quantitative belief propagation or message-passing algorithms developed by Pearl and others than the reduction-type algorithms.

Wellman (1988) provides a persuasive argument for first-order stochastic dominance (FSD) as the best formal interpretation of the informal notion of the sign of an influence, whether knowledge of A being true (high) increases or decreases belief in B being true (high). Thus a positive influence of binary variable A on variable B is defined thus:

$$I^{+}(A, B) \Longleftrightarrow P(b \mid a\ y) \geq P(b \mid \bar{a}\ y),\ \forall y \qquad [1]$$

(NB: We use the convention that uppercase letters denote variables, and lowercase their values: $\bar{a}$ means A is false.) Signs may be assigned to arcs in the network either by direct assessment or, for qualitative explanation of quantitative reasoning, as an abstraction from quantitative assessments. In the sneeze example, it is intuitively clear (and consistent with the numbers used in the example) that all influences are positive. For simplicity, we will assume the network is singly connected and limit ourselves to binary variables.

### 4.1 Three types of inference

As a prerequisite to describing the algorithm for qualitative belief propagation, we must first distinguish the three types of inference: *Predictive* (or *causal*) inference is in the same direction as the original qualitative influence. *Diagnostic* inference is in the reverse direction. *Intercausal* inference gives the qualitative impact of evidence for one variable A on another variable B, when both have influences on a third variable C, about which we have independent evidence. These three situations are illustrated below in Figure 4.

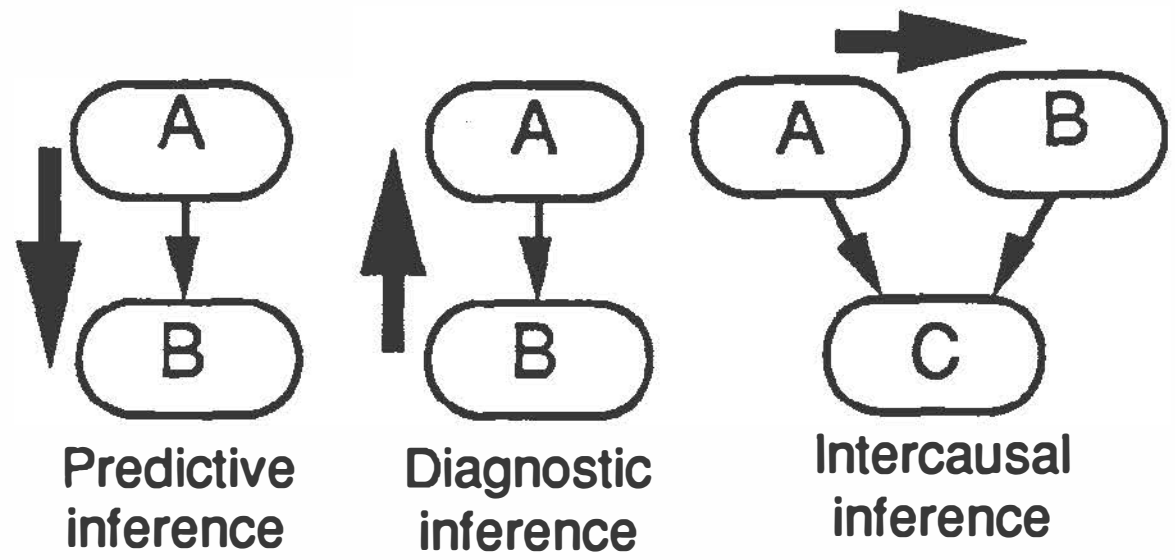

**Figure 4: Three types of inference.**

Qualitative predictive inference is quite simple. If we have positive evidence E that increases our belief in A, and the influence of A on B is positive, then E should also increase our belief in B (actually, not decrease it, given Wellman's weak definition of the direction of influence). More generally, concatenation (chaining) produces an influence whose sign is the product of the signs of its component influences.

$$I^{\delta}(E, A) \ \& \ I^{\delta}(A, B) \Rightarrow I^{\delta*\delta}(E, B)$$

Diagnostic inference is similar to predictive inference, although a little more complicated since variable A inherits any relevant predecessors of B in inverting the direction of the arrow. If we want to propagate the effect of evidence across two divergent arrows, we can simply chain the diagnostic and predictive inference. Observation of pawmarks increase belief in the cat, which in turn increases belief in the allergy.

But propagating across convergent arrows is less straightforward: If there is no diagnostic evidence for the common effect (or direct observation), e.g. for C, then the two influencing variables A and B are of course independent, and so knowledge about A has no effect on B. On the other hand, if we observe C (or have diagnostic evidence for it) A and B become dependent. Thus, intercausal inference is not a simple concatenation of predictive and diagnostic inference. While there has been much informal discussion of "explaining away" (a form of intercausal reasoning) (Henrion, 1986; Pearl, 1988), a precise characterization seems to be lacking of the general conditions under which explaining away or other qualitative intercausal reasoning applies. So we now turn to this issue.

### 4.2 Qualitative Intercausal Influence

"Explaining away" applies when A and B are two alternative causes of C, for example if the influence of A and B on C is a noisy OR. Given evidence for C, then evidence for A generally produces a reduced belief in B. But what if the influence is not a noisy-OR? What precisely are the conditions on the influence of A and B on C under which this qualitative pattern applies? Figure 5 presents the question schematically.

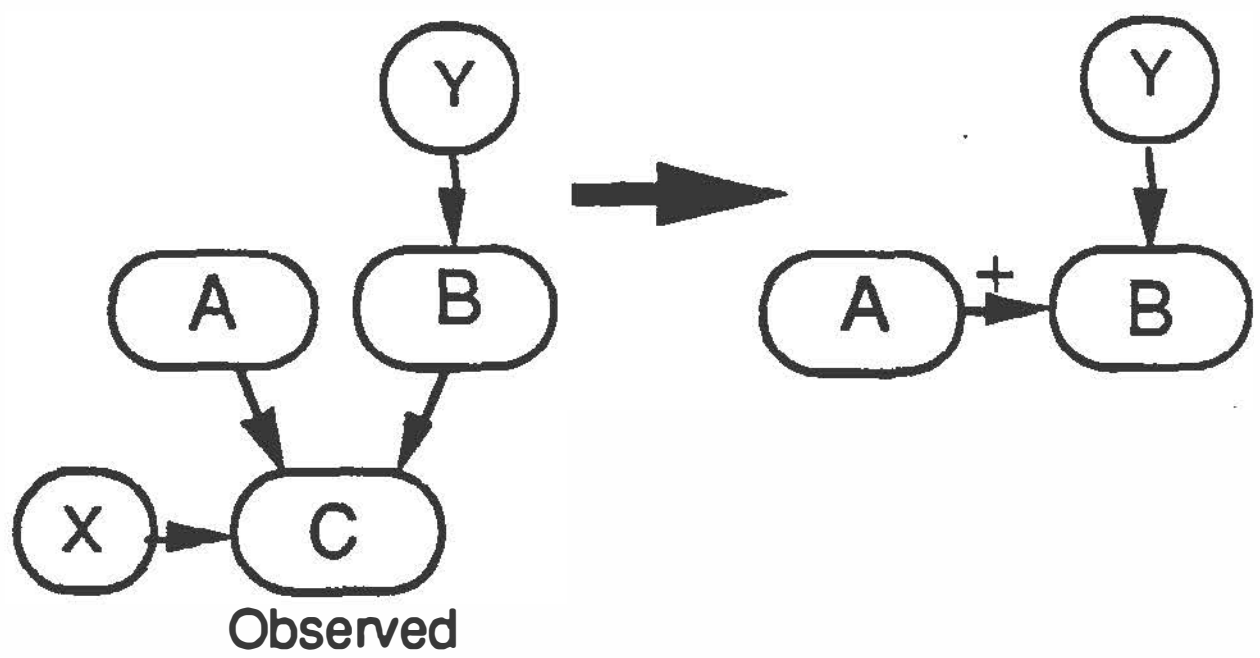

**Figure 5: Schematic of the operation of qualitative intercausal reasoning**

•**Theorem (qualitative intercausal influence):** Suppose we have three logical variables, A, B, C, all with non-zero priors, where A and B can influence C, and A and B are marginally independent (there are no paths between A and B other than through C). Suppose C is observed to be true. Then a positive qualitative influence exists between A and B, i.e.

$$P(b \mid a\, y) \geq P(b \mid \bar{a}\, y), \quad \forall y \qquad [2]$$

where y are the predecessors of B other than A, *if and only if*

$$P(c \mid a\, b\, x)\, P(c \mid \bar{a}\, \bar{b}\, x) \geq P(c \mid a\, \bar{b}\, x)\, P(c \mid \bar{a}\, b\, x), \quad \forall x \quad [3]$$

where x are the predecessors of C other than A and B. The qualitative influence from A to B is zero or negative, i.e. we replace the ≥ by = or ≤ in [1], if we replace the ≥ in condition [3] by = or ≤ respectively. •

This result is easily derived from [2] by the application of Bayes' rule. Condition [3] is reminiscent of Wellman's (1988) definition of qualitative synergy, but instead of the multiplicative form he employs the additive form (for binary variables),

$$P(c \mid a\, b\, x) + P(c \mid \bar{a}\, \bar{b}\, x) \geq P(c \mid a\, \bar{b}\, x) + P(c \mid \bar{a}\, b\, x), \quad \forall x \quad [4]$$

It can be shown that if either or both of the influences from A to C and from B to C are positive, then multiplicative synergy [3] implies additive synergy [4]. It is also easy to show that noisy-OR gates are subsynergistic for product synergy, just as Wellman showed they are subsynergistic for additive synergy. Hence, if P(C|A,B) is a noisy OR gate and C is observed present, there is a negative influence between A and B. So any further evidence for B will decrease belief in A, and vice versa, since they provide alternative explanations for C. In other words [3] is precisely the condition under which explaining away applies.

Note that if the influence has positive multiplicative synergy, A has a positive influence on B, the inverse of explaining away. For example, if the influence is a "leaky noisy AND gate", in which C has an increased chance of occurring if A and B *both* occur, then given C, knowledge of A may *increase* belief in B. For example, suppose that the presence of flammable material and an ignition source together can cause combustion, which in turn may cause smoke (which also has other possible causes). Observation of smoke can create a positive influence from flammable material to the ignition source.

### 4.3 An example of qualitative propagation

We illustrate how these ideas may be applied to provide qualitative explanation using the sneeze example, somewhat like that provided by some of our subjects. Consider question (c) from above, how does observation of barking affect our belief in the cold, given we already have observed sneezing and pawmarks? Since both the cases of convergent influences (sneezing and paw marks) are noisy ORs with observed effects, explaining away applies, that is they can be reduced to negative influences between the causes. We can generate a trace of the explanation thus:

```
Observe sneezing and paw-marks.
Impact of barking on cold?
1. Observation of barking is
evidence for dog.
2. Increased probability of dog
helps explain paw marks, and so
weakens evidence for cat.
3. Reduced probability of cat
reduces probability of allergy.
4. Reduced probability of allergy
reduces ability to explain sneezing,
and so increases probability of
cold.
In summary, observation of barking
increases probability of cold.
```

This illustrates all three kinds of propagation. Step 1 involves simple diagnostic inference over a positive influence. Step 2 involves intercausal inference, producing a negative influence from dog to cat. Step 3 involves simple predictive inference, propagating negative evidence over a positive influence. And step 4 involves intercausal inference, producing another negative influence from allergy to cold. Since there are two positive steps and two negative steps (the intercausal inferences), the chaining produces a cumulative positive influence between barking and cold, as shown in Figure 6. We should point out that the scheme as described is limited to singly connected networks of binary variables.

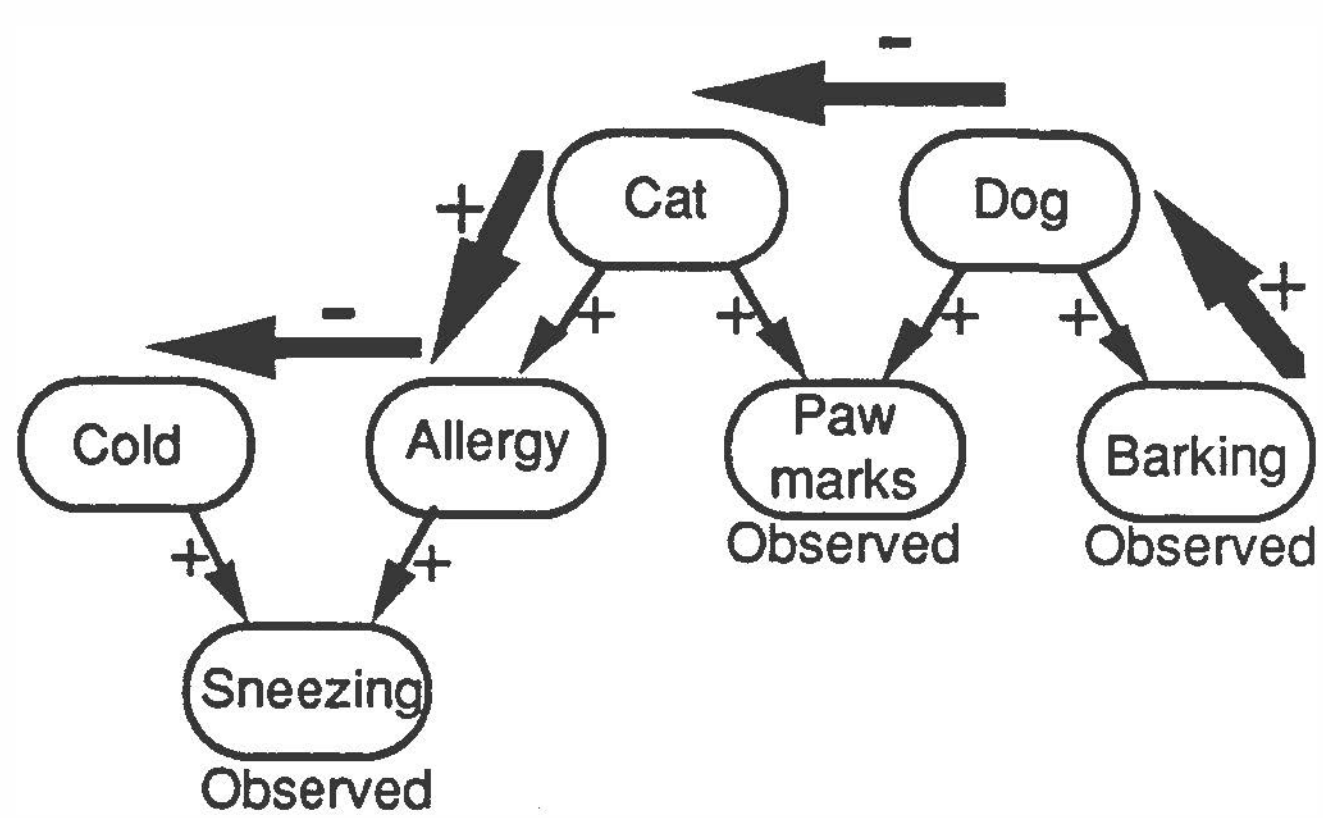

**Figure 6: Propagation of qualitative probabilistic inference.**

## 5 Scenario-based explanations

Whereas inference schemes using propagation operate on a belief network representation of knowledge, scenario-based explanations are based on scenario trees (also known as probability trees, or decision trees without the decision variables). Each path from root to an end node represents a scenario, or sequence of events.

The psychological literature suggests that it may be easier to understand scenarios if they are presented as coherent causal stories. So in an attempt to make scenarios easier to grasp, we can order the events in a scenario so that effects follow their causes, and employs causal conjunctions to link them when appropriate, for example:

```
No cold; cat causes allergy, which
causes sneezing.
Dog causes barking and paw marks; no
cat, hence no allergy; cold causes
sneezing.
```

In some scenarios, an event may deviate from what is expected, having a low probability given its predecessors. Even though a cat is present there may be no allergic reaction. In such cases, we can aid interpretation by indicating such surprises by an exception conjunction such as "but" for an event with low conditional probability:

```
No cold; cat, but no allergy, hence
no sneezing.
```

The probability of each scenario is the product of the conditional probabilities of all the events in it. Exact Bayesian inference to find the posterior probability of an event can be performed by looking at the ratio of the sum of the probabilities of all scenarios compatible with the event to the sum of all those not compatible, after eliminating all scenarios not consistent with the observations. The number of possible scenarios is generally large of course, and cognitively unmanageable. But fortunately, it is often possible to understand the essentials of what is going on by examining only a few of the most probable scenarios.

The following is a simple scenario-based explanation from the sneeze example. We ask it to explain the probability assigned to cold given sneezing has been observed:

```
why cold

Given:
  Sneezing must have been caused by
cold or allergy.

The following scenario(s) are
compatible with cold:
A. Cold and no cat hence no
   allergy                           0.47
   Other less probable
   scenario(s)                       0.06

The following scenario(s) are
incompatible with cold:
B. No Cold and cat causing
   allergy                           0.48

Scenario A is about as likely as
scenario B (0.47/0.48)
because cold in A is a great deal
less likely than no cold in B
(0.08/0.92),
although no cat in A is a great deal
more likely than cat in B (0.9/0.1).

Therefore cold is slightly more
likely than not (p=0.52).
```

QIQ first displays what is known and what can be definitely inferred from it. It then gives a list of one or more scenarios which are compatible with the target variable (cold) and a second list of scenarios which are incompatible with it. Since currently our only observation is sneezing, the variables paw marks, dog, and barking are irrelevant, and so the scenarios mention only cold, cat, and allergy.

In general there may be a vast number of possible scenarios (exponential in the number of uncertain variables), so it only gives the most probable one(s) in each list. The rest are grouped as "other less probable scenario(s)", those which collectively contribute less than 15% of the overall probability for that list. This parameter can be varied to control the length and precision of the explanation.

The next part of the explanation compares the probabilities of the most important pairs of scenarios in terms of significant differences in the probabilities of their component events. Any contrasts that are significant (probabilities differing by a factor of more than 1.2 in the default option) are mentioned in explaining the relative probabilities. The explanation lists, after "because", the contrasts favoring the more probable scenario, and then, after "although", the contrasts, if any, supporting the other scenario. This scheme is based on the principle that it is easier to judge the relative probability of two scenarios by comparing their differences than by judging their absolute probabilities. The comparisons use the relative probability phrases calibrated against numerical probability differences by Elsaesser & Henrion (1990). In this case the relatively low probability of cold in scenario A is just about balanced by the low probability of cat in scenario B. Another possible scenario which has both cold and cat (hence allergy) is not even mentioned, because having two very unlikely events its relative probability is negligible relative to the two scenarios each with a single unlikely event. So, it is contained in the "other less probable scenario(s)" group.

A second example explanation assumes that paw marks and barking as well as sneezing have been observed:

```
why cold

Given:
  Sneezing must have been caused by
cold or allergy.
  Paw Marks could have been caused by
  cat or dog or another unknown cause.
  Barking must have been caused by
dog.

The following scenario(s) are
compatible with cold:
A. Cold and no cat hence no
   allergy and dog                   0.38
B. Cold and cat causing allergy
   and dog                           0.05
   Other less probable scenario(s)   0.01
                                     ----
   Total probability of cold         0.44

The following scenario(s) are
incompatible with cold:
C. No Cold and cat causing
   allergy and dog                   0.56

Scenario A is much more likely than
scenario B (0.38/0.05)
because no cat in A is a great deal
more likely than cat in B
(0.90/0.10).

Scenario A is somewhat less likely
than scenario C (0.38/0.56)
because cold in A is a great deal
less likely than no cold in C
(0.08/0.92),
although no cat in A is a great deal
more likely than cat in C
(0.90/0.10).

Therefore cold is fairly unlikely
(p=0.44).
```

Note that several techniques are provided to abstract and simplify the explanation. First, only relevant events are considered, that is events whose consideration affects the target probability given available observations. Second, only those scenarios that contribute more than 10% of the probability to the target event (or its complement) are listed explicitly. This can drastically simplify the explanation, since many real cases seem to be like the example above, where a few (two to four) scenarios turn out to have the bulk of the probability, and the vast mass can be ignored without significant error. Thirdly, in comparisons of the relative probability of pairs of scenarios, only those events with substantially different probabilities are mentioned.

Additional abstraction techniques could provide further simplification. Some linked variables might be combined so that they can be treated as one. For example, allergy might be combined into cat, considering cat to cause sneezing directly. This reduction of variables can reduce the number of distinct scenarios and also the complexity of each scenario. Another improvement in a decision context would be to consider the importance of a scenario in terms of expected utility rather than simply probability. In a medical context, low probability scenarios leading to death may have a stronger claim to be listed explicitly than higher probability scenarios with less interesting consequences.

There is considerable psychological evidence for the prevalence of scenario-based reasoning in human thinking, and of the importance of coherent, causal stories (e.g. Pennington & Hastie, 1988). The psychological literature has focussed generally on the ways in which this leads to systematic distortions and biases in probabilistic judgment. Our approach here is to generate explanations matched to human preferences for explanatory stories, but to select and present the scenarios in such a way that they provide a reasonable guide to the relative probabilities of interest.

## 6 Conclusions

We have outlined a variety of approaches to explaining probabilistic knowledge, including combinations of graphic, textual, and numeric information, and two new schemes for explaining probabilistic reasoning. Any specific explanation will be too verbose for some, too brief for others, and simply confusing for yet others. The art in designing a good explanation is to match the style and focus to the skills and interests of the person to whom you are explaining. Users are generally the best judge of their preferences and interests, and so it is important to provide them with levers to control the style and completeness, such as the probability cut-off parameters mentioned in scenario generation. Providing both phrases and numbers should please most people, and may even give numerophobes the opportunity to become familiar with the relationships.

Qualitative belief propagation and scenario-based explanations will be appealing in different situations. Propagation seems to work well in cases with singly connected networks and strong qualitative influences. The analysis of qualitative intercausal reasoning we have presented provides a principled basis for understanding some important patterns of intuitive reasoning, as well as extending formal methods for qualitative probabilistic reasoning. While belief propagation provides intuition into the direction of the impacts of evidence, in ts purely qualitative form it provides little guidance about the magnitude of effects or probabilities. Scenario-based explanations can work with multiply connected networks. Scenario-based reasoning also appears to offer a natural way of reconciling probabilistic reasoning to possible-worlds logic-based approaches to reasoning. Both schemes lend themselves to convenient abstraction and simplification, which are essential in generating comprehensible explanations.

This paper summarizes work from an initial exploratory phase of our research program into explaining probabilistic reasoning. There is considerable scope for developing and refining of these techniques. Initial experimental evaluation suggests the value of such explanations in improved user insight (Druzdzel, 1990). But a more definitive understanding of the merits of such schemes must awa t more extended empirical comparisons of their effectiveness under a variety of conditions.

## Acknowledgements

We are most grateful to Michael Wellman for insightful comments.